\documentclass[lettersize,journal]{IEEEtran}
\usepackage{amsmath,amsfonts}
\usepackage{amssymb}
\usepackage{algorithmic}
\usepackage{array}
\usepackage[caption=false,font=normalsize,labelfont=sf,textfont=sf]{subfig}
\usepackage{textcomp}
\usepackage{stfloats}
\usepackage{url}
\usepackage{subcaption}
\usepackage{pifont}
\usepackage{verbatim}
\usepackage{graphicx}
\usepackage{cite}
\usepackage{ragged2e}
\usepackage{rotating}
\usepackage{color}
\usepackage{xcolor}
\usepackage{float}
\usepackage{colortbl}
\usepackage{booktabs}
\usepackage{multirow}
\usepackage{hyperref}

\newcommand{\red}[1]{\textcolor{red}{#1}}

\newcommand{\green}[1]{\textcolor[RGB]{26,217,22}{#1}}
\newcommand{\modify}[1]{\textcolor{black}{#1}}

\usepackage[ruled,vlined]{algorithm2e}
\IEEEpubid{\begin{minipage}{\textwidth}\ \centering
        Copyright \copyright 20xx IEEE. Personal use of this material is permitted. \\ However, permission to use this material for any other purposes must be obtained from the IEEE by sending an email to pubs-permissions@ieee.org.
\end{minipage}}

\begin{document}

\title{LTCA: Long-range Temporal Context Attention for Referring Video Object Segmentation}

\author{
Cilin Yan$^{*}$, Jingyun Wang$^{*}$, Guoliang Kang$^{\dagger}$

\thanks{$^{*}$Equal contribution. $^{\dagger}$Corresponding author.} 
\thanks{Cilin Yan, Jingyun Wang and Guoliang Kang are with School of Automation Science and Electrical Engineering, Beihang University, Beijing 100191, China~(e-mail: clyan@buaa.edu.cn, 19231136@buaa.edu.cn, kgl.prml@gmail.com)}
}

\markboth{Journal of \LaTeX\ Class Files,~Vol.~14, No.~8, August~2021}%
{Shell \MakeLowercase{\textit{et al.}}: A Sample Article Using IEEEtran.cls for IEEE Journals}

\IEEEpubidadjcol

\maketitle

\begin{abstract}

Referring Video Segmentation (RVOS) aims to segment objects in videos given linguistic expressions. 
The key to solving RVOS is to extract long-range temporal context information from the interactions of expressions and videos to depict the dynamic attributes of each object.
Previous works either adopt attention across all the frames or stack dense local attention to achieve a global view of temporal context.
However, they fail to strike a good balance between locality and globality, and the computation complexity significantly increases with the increase of video length. 
In this paper, we propose an effective long-range temporal context attention (LTCA) mechanism to aggregate global context information into object features.  
Specifically, we aggregate the global context information from two aspects. 
Firstly, we stack sparse local attentions to balance the locality and globality.
We design a dilated window attention across frames to aggregate local context information
and perform such attention in a stack of layers to enable a global view.
Further, we enable each query to attend to a small group of keys randomly selected from a global pool to enhance the globality. 
Secondly, we design a global query to interact with all the other queries to directly encode the global context information. 
Experiments show our method achieves new state-of-the-art on \modify{four} referring video segmentation benchmarks. 
Notably, our method shows an improvement of $\mathbf{11.3\%}$ and $\mathbf{8.3\%}$ on the MeViS $val^u$ and $val$ datasets respectively.
The code is available at \url{https://github.com/cilinyan/LTCA}.

\end{abstract}

%
\section{Introduction}

\begin{figure}[!t]
  \centering
  \includegraphics[width=1.0\linewidth]{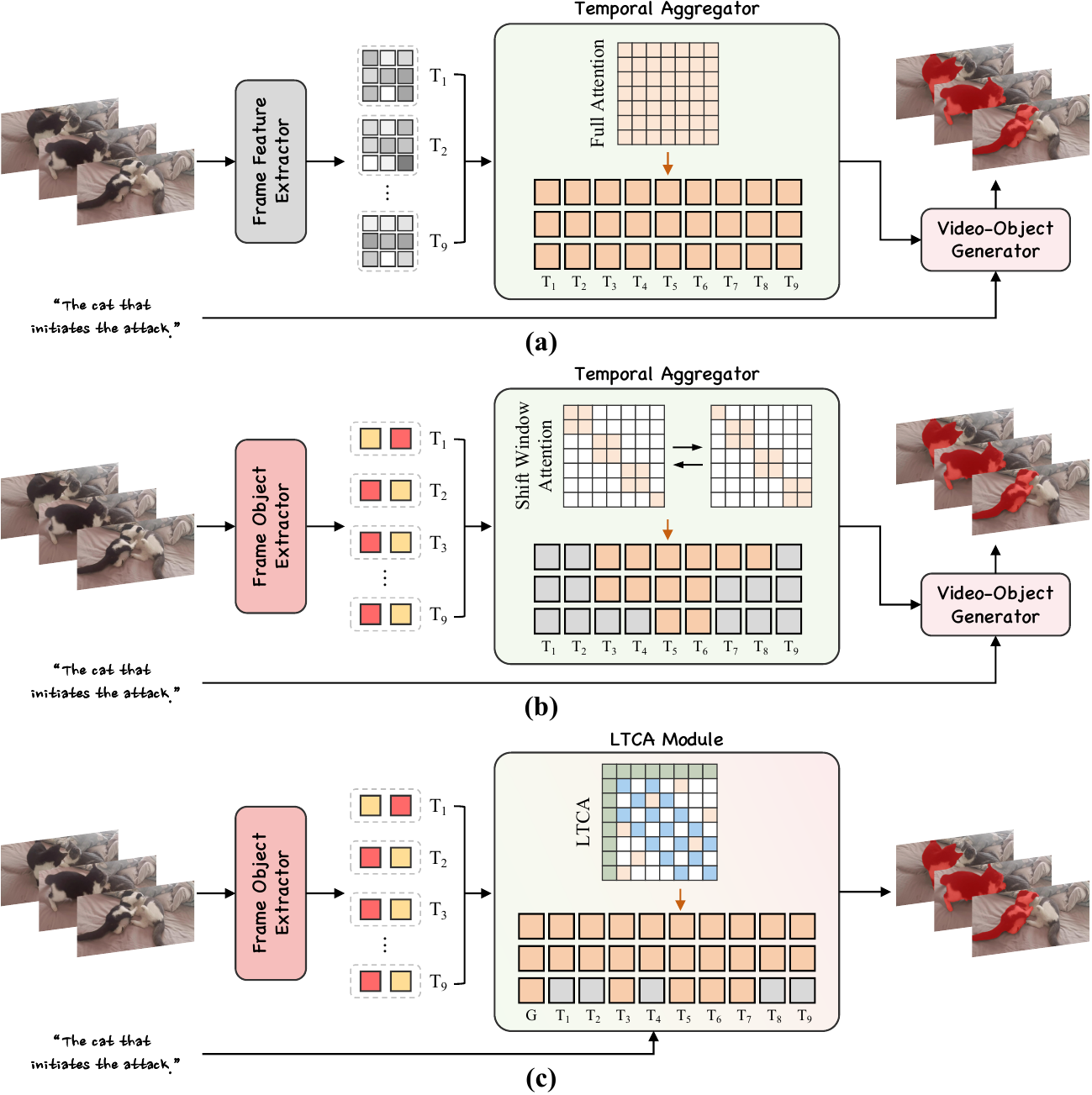}
  \vspace{-5mm}
\caption{
Comparison of current query-based referring video segmentation~(RVOS) pipelines. 
(a) Modeling frame feature sequence using full attention,
(b) Modeling frame object sequence using shift window attention,
(c) Modeling frame object sequence using LTCA.
}

\vspace{-6mm}
\label{fig:cpr_lmpm_zoe}
\end{figure}

\modify{
Referring video object segmentation~(RVOS)~\cite{botach2022end,wu2022language} is a specialized form of segmentation task~\cite{li2023adversarial,li2017towards,zhao2018understanding,zhao2017self,chen2021building,chen2024rsprompter,wang2024learn,yan2024piclick,wang2024rsbuilding,wang2024recliplearnrectifybias,chen2024ov,wang2022fully,zhu2012fast} that requires identifying and segmenting target objects in a video using detailed language expressions as guidance.
}
RVOS is challenging as we need to build interactions between expressions and videos and 
align context information across modalities to localize and segment the desired object.
Though challenging, RVOS is of vital importance in practice, as it is natural to depict our demand in language expressions and segment the desired objects in videos under the guidance of natural language.
RVOS may benefit many real-world applications, \emph{e.g.,} \modify{autonomous driving~\cite{wu2023referring,wu2023language}, video surveillance~\cite{yang2021hierarchical,vezzani2013people}, video editing~\cite{khoreva2018video,mei2024slvp}, human-robot interaction~\cite{gross2016multi,guadarrama2013grounding}, \emph{etc}. 
For instance, in human-robot interaction, through RVOS, users can instruct a robot to complete a specific task, which is important for seamless interaction and task completion in dynamic environments.}

\modify{
The key to solving RVOS lies in extracting long-range temporal context information from the interactions between expressions and videos, and effectively capturing the dynamic attributes of each object.
Among the long-range temporal context information, there exist both local and global information.
For locality, we expect the model to emphasize local context. To realize this, each frame only need to attend to its neighbors within a certain temporal range to extract the context information. For globality, we expect the model to capture global temporal information across all frames in the video.
}
\IEEEpubidadjcol
Currently, there exist two typical ways to aggregate the long-range temporal context information in RVOS. 
One way (\emph{e.g.,}~\cite{wu2022language} in Fig.~\ref{fig:cpr_lmpm_zoe}(a)) is to adopt a full attention across all the frames, \emph{i.e.,} 
each frame is capable of attending to all the other frames to aggregate useful information. 
However, for long sequences of frames, the magnitudes of full attention weights may become quite close, 
rendering it hard to emphasize local important context information. 
As a result, the full-attention way may capture a global view of temporal context, but sacrifice important local information, exhibiting a bad locality.
Another way is to apply dense local attention (\emph{e.g.,} shift window attention~\cite{ding2023mevis} in Fig.~\ref{fig:cpr_lmpm_zoe}(b)) to aggregate local context information and stack such attention layers to obtain a global context.
However, as the length of video increases, in order to capture global temporal information, the number of attention layers should also increase, 
resulting in larger time and memory costs.
Thus, in this way, although the locality is enhanced, the global context modeling ability is weakened. 
To summarize, previous works fail to strike a good balance between locality and globality, and the computation complexity may significantly increase with the increase of video length. 
\modify{
Various attention mechanisms have been explored in action/motion recognition task ~\cite{shu2021spatiotemporal,tang2019coherence,shu2019hierarchical,yan2020higcin,shu2022multi,shu2020host}. 
However, in RVOS task, we need to model the long-range temporal context information for each spatial position and model the text-video interaction, the massive attention computation of which 
renders the attention mechanism proposed in previous action/motion recognition task  not applicable to the RVOS task.
}

In this paper, we introduce an effective long-range temporal context attention (LTCA) mechanism for RVOS task.
Specifically, the queries, which are initialized by the linguistic expression embeddings, interact with extracted frame features to encode the object and context information for each frame. 
\modify{LTCA (as illustrated in Fig.~\ref{fig:cpr_lmpm_zoe}(c)) then aggregates the long-range temporal context information and manages to achieve a good balance between locality and globality with the following specific designs:}
\modify{Firstly, we design a dilated window attention mechanism across frames to aggregate local context information effectively and perform this attention over multiple layers to enable a comprehensive global view.}
\modify{Secondly, we design a random attention mechanism to enhance overall globality by allowing each query to attend to a small, randomly selected group of keys from a global pool.}
\modify{Thirdly, we introduce global queries to interact with all the other queries to encode the global context information and further improve the globality. }
Thanks to LTCA, we may directly compare aggregated global query features and the frame feature maps to generate the object masks across frames,
without the need to rely on a video-object generator to correlate expressions and aggregated features (as shown in Fig.~\ref{fig:cpr_lmpm_zoe}(a-b)).

Extensive experiments are conducted on representative benchmarks including MeViS~\cite{ding2023mevis}, Ref-YouTube-VOS~\cite{seo2020urvos}, Ref-DAVIS17~\cite{khoreva2019video} and \modify{A2D-Sentences~\cite{gavrilyuk2018actor}}. 
Our method achieves state-of-the-art performance on \modify{four} benchmarks. 
Note that among these benchmarks, MeViS contains more motion expressions, which proposes a higher demand for utilizing temporal information.
Our superior performance on MeViS (outperforming previous state-of-the-art by 11.3\%) 
verifies the effectiveness of LTCA to aggregate long-range temporal information. 

In summary, our contributions are as follows:

\begin{itemize}
    \item
    We propose a new attention mechanism LTCA to capture the long-range temporal context for RVOS. LTCA stacks sparse local attention (including dilated window attention and random attention) layers to balance locality and globality, and directly aggregates global information via specifically designed global queries in global attention.

    \item Based on LTCA, we propose a simplified framework, where queries are firstly encoded through interactions between expressions and frame features, then object queries aggregate temporal context information through LTCA, and finally the object masks across frames are generated by comparing global queries with the 
    frame features. 
    
    \item Extensive experiments on \modify{four} representative benchmarks (\emph{i.e.,} MeViS, Ref-YouTube-VOS, Ref-DAVIS17, \modify{A2D-Setences}) show that our method outperforms previous works remarkably, achieving new state-of-the-art. 
\end{itemize}
\begin{figure*}[!t]
  \centering
  \includegraphics[width=1.0\linewidth]{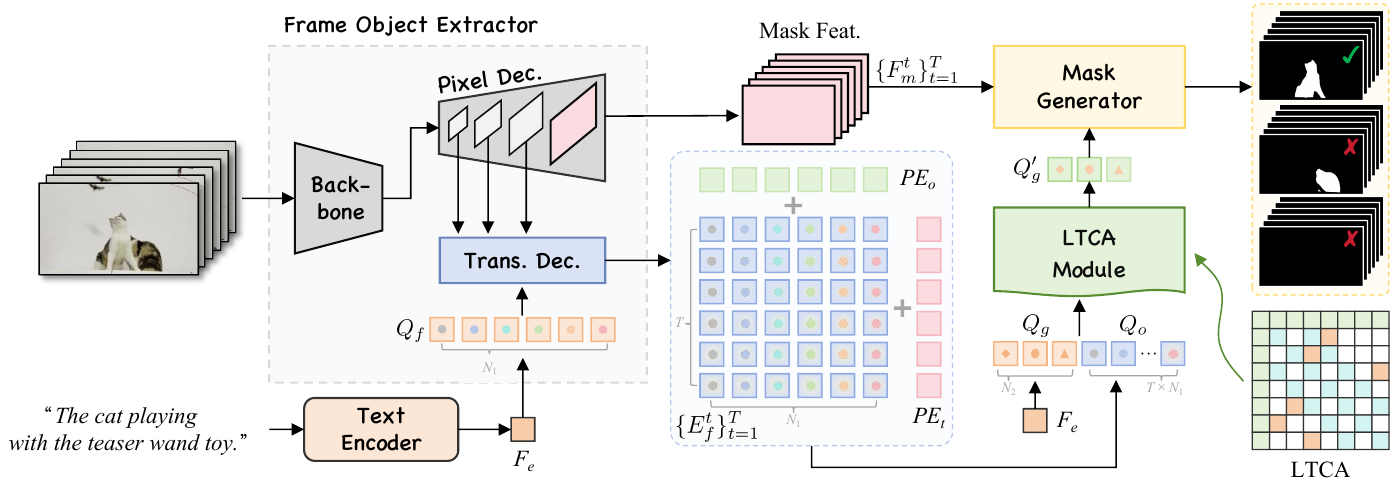}
  \vspace{-3mm}
\caption{
\textbf{The overview architecture of our method. }
First, all input frames are fed to a transformer-based extractor to generate object-centric embeddings $\{E_{f}^t\}_{t=1}^T$.
Then we flatten $\{E_{f}^t\}_{t=1}^T$ as frame object queries $Q_o$ and concatenate it with a set of learnable global queries $Q_g$, which is initialized with text embedding of given linguistic expressions.
Then the concatenated queries are fed to an LTCA module to conduct efficient information interaction among frame object queries $Q_o$ and linguistic-aware global queries $Q_g$. 
The output global queries $\widetilde{Q}_g$ are adopted to generate segmentation mask sequences of target objects.
}
\vspace{-3mm}
\label{fig:architecture}
\end{figure*}

\section{Related Work}

In this section, we first introduce the related works of video instance segmentation. Afterward, we give a brief review of referring image segmentation. Finally, we introduce referring video object segmentation.

\subsection{Video Instance Segmentation}
Video instance segmentation~(VIS)~\cite{huang2022minvis,jiang2022stc,wu2022idol,MaskTrack-R-CNN,yang2021crossvis,yan2022solve,fang2024learning,wang2024ov,zhang2023dvis,zhang2023dvis++} is a joint vision task, which requires to identify, segment and track the interested instances in a video simultaneously. 
Existing VIS methods can be mainly divided into two categories: online and offline approaches. 
Online methods~\cite{huang2022minvis,jiang2022stc,wu2022idol,MaskTrack-R-CNN,yang2021crossvis} process a video frame-by-frame, segmenting objects in a frame level and associating instances across time. 
MaskTrack R-CNN~\cite{MaskTrack-R-CNN} incorporates a tracking head based on Mask R-CNN~\cite{he2017maskrcnn} and associates instances in different frames by heuristic cues.
Following the recent success of Transformer~\cite{vaswani2017attention}, several Transformer-based methodologies have been introduced. 
For instance, MinVIS~\cite{huang2022minvis} recognizes the discriminative nature of query embeddings and employs a straightforward strategy by exclusively aligning these embeddings between the current and adjacent frames. 
Meanwhile, CTVIS~\cite{yang2021crossvis} learns more robust and discriminative instance embeddings by aligning the training and inference pipelines.
Offline methods~\cite{hwang2021video,wu2022seqformer,heo2022vita} take the entire video as input, performing segmentation on all frames and associating instances at one time.
SeqFormer~\cite{wu2022seqformer} processes spatio-temporal features and directly predicts instance mask sequences, while VITA~\cite{heo2022vita} proposes to associate frame-level object tokens without using dense spatio-temporal backbone features and reduce the memory consumption on long videos.

\subsection{Referring Image Segmentation}

Referring Image Segmentation~(RIS)~\cite{shang2022cross,li2023fully,wu2024f} aims to segment the target object in the given static image referred by language.
Early studies~\cite{hu2016segmentation,li2018referring} have adhered to a concatenate-then-convolve approach, in which the fusion of language and image features is accomplished through concatenation.
Subsequent works~\cite{chen2019see,han2021dynamic} have enhanced segmentation performance by employing RNN or dynamic convolution techniques, or by investigating optimal position for executing language-image fusion.
Building on the success of attention architecture~\cite{dosovitskiy2020image}, current studies~\cite{shi2018key,feng2021bidirectional,yang2022lavt} primarily employ cross-attention modules for language-image fusion. 
For instance, LAVT~\cite{yang2022lavt} integrates the fusion of visual and linguistic features using unidirectional cross-attention at the intermediate layers of encoders.
Additionally, VPD~\cite{zhao2023unleashing} attempts to harness semantic information in diffusion models for RIS, while LISA~\cite{lai2023lisa} and PixelLM~\cite{ren2023pixellm} draws on the language generation capabilities of multimodel large language models~\cite{liu2023llava,fang2023instructseq} to produce a segmentation mask in response to a complex and implicit query text.
SESAME~\cite{wu2023see} introduces supplementary data to resolve the problem where LISA may be misled by false premises when queries imply the existence of objects that are not actually present in the image.

\subsection{Referring Video Object Segmentation}

Referring Video Object Segmentation (RVOS) is a challenging multi-modal video task, which aims to segment the target object
indicated by a given language expression across the entire video. 
Most existing benchmarks of RVOS, including Ref-YouTube-VOS~\cite{seo2020urvos}, Ref-DAVIS 2017~\cite{khoreva2019video} and A2D-Sentences~\cite{gavrilyuk2018actor}, primarily focus on salient objects, utilizing language descriptions of static attributes. 
MeViS~\cite{ding2023mevis} is a large-scale video dataset that contains videos with longer duration and expressions referring to an object with highly dynamic features, such as movements and state changes. 
A robust capability for long-range temporal feature modeling is essential for tackling MeViS dataset, but most existing methods~\cite{seo2020urvos,chen2022multi,wu2022language,botach2022end,ding2023mevis} lack this capability.
MTTR~\cite{botach2022end} adopts a query-based end-to-end framework to decode objects from multi-modal features frame by frame independently, thus failing to model the inter-frame temporal information.
ReferFormer~\cite{wu2022language} employs full attention on frame-level tokens to model temporal information.
However, the overly long frame features and quadratic complexity of full attention results in vast time and memory costs.
Moreover, for input sequences of long-duration videos, the distribution of attention weight becomes smooth, 
making the model only focus on global context information but fail to effectively model the local temporal information.
R$^2$VOS~\cite{li2022r} began by augmenting the dataset with mismatched video-text pairs. It then utilized a contrastive learning approach to identify and eliminate these mismatches, thereby refining the model's grasp of language features.
TempCD~\cite{tang2023temporal} aligned the natural language expression with the objects' motions and their dynamic associations at the global video level and segmented objects at the frame level.
\modify{
DMFormer~\cite{gao2023decoupling} enhances explicit and discriminative feature embedding and alignment by introducing lightweight subject-aware and context-aware branches.
HTR~\cite{miao2024temporally} addresses the challenge of temporal instance consistency by introducing a hybrid memory framework with inter-frame collaboration and a Mask Consistency Score metric.
Sun et al.~\cite{sun2023unified} propose a unified framework for multi-modality VOS, employing self-attention to process inputs as tokens and using reinforcement learning to optimize filter networks for highlighting significant information.
}
However, \cite{tang2023temporal} and \cite{luo2024soc} utilize the full attention mechanism, with the computational complexity of $O(n^2)$, rendering them impractical for long video scenarios.
Recent works~\cite{ding2023mevis,yan2023referred} exploits a shift-window attention mechanism in the Transformer architecture, motivated by VITA~\cite{heo2022vita}. 
Although the shift-window attention reduces the inference cost, it only models the local context information, losing the ability to model global context information for long videos.
In this study, we introduce an efficient long-range temporal context attention mechanism that strikes a balance between locality and globality, while simultaneously simplifying the framework of RVOS.

\begin{figure*}[!t]
  \centering
  \includegraphics[width=1.0\linewidth]{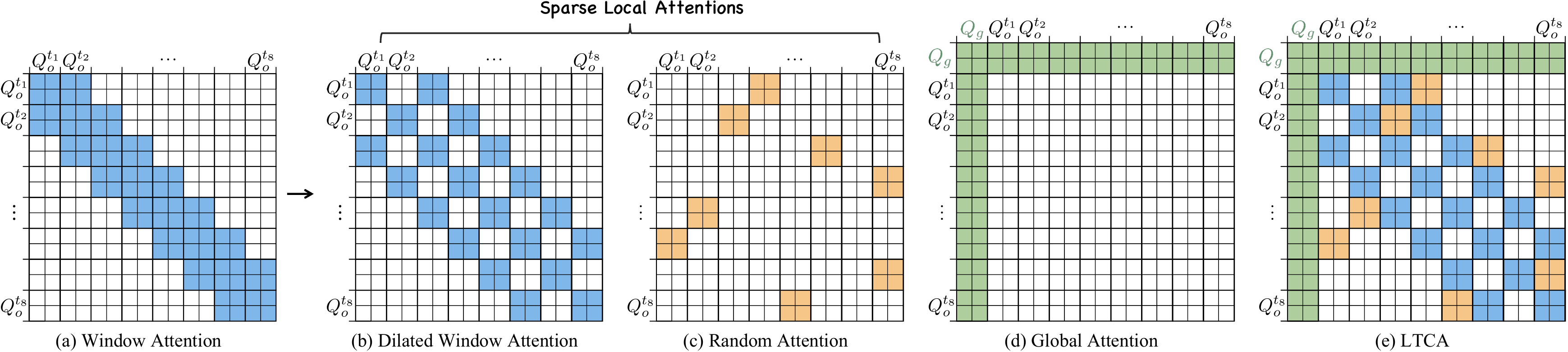}
  \vspace{-3mm}
\caption{
Visualization of different attention patterns.
For ease of visualization, we set $N_1=2$, $N_2=2$, $w = 3$, $d = 2$ and $r = 1$.
}
\vspace{-3mm}
\label{fig:cpr_different_attn}
\end{figure*}

\section{Method}

In this work, we propose a new method for Referring Video Segmentation task.
RVOS aims to segment target objects in all frames of the video according to a given linguistic reference. 
The visual input is a sequence of video frames and the textual input is a sentence describing specific objects.

The general framework of our method is illustrated in Fig.~\ref{fig:architecture}. 
Given a language expression containing $L$ words as text input, we encode it in word level by an off-the-shelf text encoder of RoBERTa~\cite{liu2019roberta}. 
We then obtain a sentence-level text feature $F_e$ by averaging the obtained word embeddings.
Object queries $Q_{f}$ are initialized by repeating $F_e$ for $N_1$ (number of potential objects per frame) times and global queries $Q_{g}$ are initialized by repeating $F_e$ for $N_2$ (total number of potential objects in the whole video) times.
Passing the video frames $V=\{I^t\}_{t=1}^T$ and object queries $\{Q_{f}^t\}_{t=1}^T $ to the Frame Object Extractor $\mathcal{E}_{f}$ (Sec. ~\ref{sec:frame detector}), mask features $\{ F_m^t \}_{t=1}^T $ and embeddings of $N_1$ potential objects $\{E_{f}^t\}_{t=1}^T$ in each frame are generated.
We then flatten the object embeddings $\{E_{f}^t\}_{t=1}^T$ as frame object queries $Q_{o}$ and concatenate them with the linguistic-aware global queries $Q_{g}$ to form the input of the LTCA (Long-range Temporal Context Attention) Module $\mathcal{E}_v$. 
Inside the LTCA Module, we propose a new LTCA mechanism (Sec.~\ref{sec:Attention_Mechanism}), consisting of Global Attention and Sparse Local Attention, to effectively aggregate global context information into object features and keep a good balance between globality and locality.
Therefore, each query in $\widetilde{Q}_g$ processed by LTCA module represents an object instance proposal within the video regarding the linguistic expression. 
Sending these global queries $\widetilde{Q}_g$ to Mask Generator (Sec.~\ref{sec:Mask_Generation}), we finally generate a sequence of target masks $\{M_i\}_{i=1}^{N_2}$.

\subsection{Baseline: Segment via Shift Window Attention}
\label{sec:Brief_Overview_of_LMPM}
LMPM~\cite{ding2023mevis} performs RVOS task by aggregating temporal information across the entire video via shift-window attention mechanism. 
LMPM obtains mask features $\{ F_m^t \}_{t=1}^T $ and object embeddings $\{E_{f}^t\}_{t=1}^T$ from Frame Object Extractor. 
Flattening frame-level object queries $\{E_{f}^t\}_{t=1}^T$ as frame object queries $Q_{o}$, LMPM passes it to the temporal aggregator with shifted window attention mechanism to obtain contextual temporal information $\hat{Q}_{o}$ across $T$ frames.
Subsequently, the Transformer decoder treats the linguistic-aware global queries $Q_{g}$ as queries and uses the obtained object embeddings $\hat{Q}_{o}$ as keys and values, to predict the masks and confidence scores of the target objects.

Although LMPM enhances the locality, the global context modeling is weakened.
For example, given the length of window $w_s$ and the number of layer $k$ in shift-window attention mechanism, two frames can contact with each other only if the distance between them is less than $\Theta (kw_s / 2)$. 
In other words, as the length of the target video increases, in order to capture global temporal information of video, the number of attention layers should also increase, thus leading to larger time and memory costs.
Besides, the last Transformer decoder is unnecessary since the frame-level object queries $Q_{o}$ can be simply encoded into linguistic-aware global queries $Q_{g}$ in the object encoder.
Based on the observations above, we propose a new framework with a newly designed LTCA module in the following sections.

\subsection{Frame Object Extractor}
\label{sec:frame detector}
Mask2Former~\cite{cheng2022masked} is adopted here as a frame object Extractor $\mathcal{E}_{f}$. 
A sequence of $T$ video frames $V=\{I^t\}_{t=1}^T$ are sent into the extractor as visual input and frame-level low-resolution image features are extracted by the visual backbone. 
A pixel decoder gradually upsamples these low-resolution features to form multi-scale image features. 
For one branch, the final high-resolution per-pixel image features are treated as mask features 
$\{F_M^t\}_{t=1}^T$
for further mask generation.
For the other branch, image features of different scales are sent into a Transformer decoder as keys and values for different layers successively and are decoded through cross-attention with the object queries 
$\{Q_f^t\}_{t=1}^T$
, yielding object embeddings $\{E_{f}^t\}_{t=1}^T$.
The process can be written as follows,
\begin{equation}
    \{F_M^t\}_{t=1}^T, \{E_{f}^t\}_{t=1}^T = \mathcal{E}_{f} (\{I^t\}_{t=1}^T).
\end{equation}
As a result, each potential object in each frame is represented by an output object query.
Then, the output frame-level object queries are used to obtain global temporal information in the Long-range Temporal Context Attention mechanism and generate linguistic-guided segmentation kernels in Mask Generator.

\subsection{Long-range Temporal Context Attention~(LTCA) Module}
\label{sec:Attention_Mechanism}

\subsubsection{\textbf{LTCA Module}}

Long-range Temporal Context Attention (LTCA) Module $\mathcal{E}_{v}$ simultaneously aggregates long-range temporal context information and models video-level object representations with linear computational complexity.
LTCA Module consists of a stack of our proposed Transformer encoders as modified versions of the standard ones.
The key components of our Transformer encoder include:
a) global queries, which extract video-level object representations,
b) masked attention operator.
\footnote{For explanatory purposes, we use an attention mask to illustrate the Long-range Temporal Context Attention mechanism. We achieve linear complexity computation by performing roll operations on keys.}, which models videos with linear computational complexity.

We flatten the object embeddings $\{E_{f}^t\}_{t=1}^T \in \mathbb{R}^{T\times N_1 \times D}$ as frame object queries $Q{'}_{o} \in \mathbb{R}^{T  N_1 \times D}$, where $D$ is the feature dimension of object embeddings.
We then combine learnable object position embeddings $PE_{o}\in \mathbb{R}^{N_{1} \times D}$ with non-learnable sinusoidal frame positional embeddings $PE_{t} \in \mathbb{R}^{T\times D}$, and then add them to frame object queries $Q{'}_{o}$ to obtain $Q_{o} \in \mathbb{R}^{T  N_1 \times D}$ as input for the encoder, which can be represented as:

\begin{equation}
    Q_{o}^{(i - 1) \times N_1 + j} = Q_{o}^{'(i - 1) \times N_1 + j} + PE_{o}^{j} + PE_{t}^{i},
\end{equation}
where $i \in \{1, 2, ..., T\}$ and $j \in \{ 1, 2, ..., N_1\}$.
We then concatenate the frame object queries $Q_{o}$ with linguistically-aware global queries $Q_{g} \in \mathbb{R}^{N_2 \times D}$ to form the input $Q_{v}$ for the LTCA Module $\mathcal{E}_{v}$, illustrated as follows:
\begin{equation}
    Q_{v} = [Q_{g}, Q_{o}] \in \mathbb{R} ^{N_2 + TN_1} .
\end{equation}
\modify{The attention operation of LTCA module can be expressed as}
\begin{equation}
    Q_{v}^{l} = \mathrm{softmax}(M^{l-1} + Q^{l}K^{l\,T})V^{l} + Q_{v}^{l-1}.
\label{eq:mask_attn}
\end{equation}

\noindent
Here, $l$ is the layer index of LTCA module, $Q_{v}^{l}$ refers to $N_2 + TN_1$ $D$-dimensional query features at the $l^{\mathrm{th}}$ layer and $Q^l,K^l,V^l\in \mathbb{R}^{(N_2 + TN_1) \times D}$ are the query features under linear transformation $f_Q(\cdot)$, $f_K(\cdot)$ and $f_V(\cdot)$ respectively.
$Q_{v}^{0}$ is the input query features $Q_{v}$.
Moreover, $M^{l-1}$ denote the attention mask for the $(l-1)^{\mathrm{th}}$  layer.
As illustrated in Fig.~\ref{fig:cpr_different_attn}(e), the positions that are non-white in $M^{l-1}$ are filled with zeros, while all other colored positions are filled with $-\infty$. 
The acquisition of $M^{l-1}$ will be described in Sec.~\ref{sec:LTCA_mechanism}.
Through LTCA module, we can derive video-level query representations, denoted as $\widetilde{Q}_{g}$, which can be written as, 
\begin{equation}
    [\widetilde{Q}_{g}, \widetilde{Q}_{o}] = Q_{v}^{-1} = \mathcal{E}_{v}(Q_{v}) = \mathcal{E}_{v}([Q_{g}, Q_{o}]).
\end{equation}

\subsubsection{\textbf{LTCA mechanism}}
\label{sec:LTCA_mechanism}
\modify{Our LTCA module uses Long-range Temporal Context Attention~(LTCA) to capture long-range temporal context for RVOS.}
LTCA is designed to aggregate temporal context information in two ways: one is to balance the locality and globality via stacking sparse local attention layers, and the other one is to directly aggregate the global information.

\noindent
\textbf{Sparse Local Attention}
Firstly, we introduce window attention across frames to aggregate local context information.
As shown in Fig.~\ref{fig:cpr_different_attn}(a), the embedding of the current frame can attend to the embeddings of its $w$ neighboring frames on both sides, which is beneficial for the construction of local contextual temporal information.
\modify{
As shown in Eq.~\ref{eq:mask_attn}, the forms of different attention operations differ only in the definition of attention mask $M$.
With $\phi(i)$ representing the index of the video frame where query $i$ is extracted, the window attention mask $M_{ij}$ with respect to query $i$ is defined as
\begin{align} 
M_{ij} = \begin{cases}
0, & \text{if } |\phi(i) - \phi(j)| \leq \dfrac{w}{2} \\
-\infty, & \text{otherwise}
\end{cases}.
\end{align}
Note that among all mask definitions of local attention, $i$ and $j$ are not in the global query set. 
From Eq.~\ref{eq:mask_attn}, $M_{ij} = 0$ means query $i$ can attend to query $j$ normally. 
In contrast, when $M_{ij} = -\infty$, query $i$ cannot attend to query $j$ as the corresponding attention weight approximates zero.
}
However, the receptive field of window attention, which is a form of dense local attention, is limited and fails to strike a good balance between locality and globality. Therefore, we further introduce dilated window attention to address this issue inspired by dilated convolutions~\cite{yu2015multi}.
As shown in Fig.~\ref{fig:cpr_different_attn}(b), dilated window attention is an extension of window attention, where the current frame attends to its neighboring frames at an interval of $d$ frames within a certain window range. Window attention can also be viewed as dilated window attention with $d=1$.
The computational complexity of dilated window attention is $O(w  N_1  T)$, which is independent of $d$.
The introduction of dilated window attention, a form of sparse local attention, effectively expands the receptive field of each frame, thereby enabling better aggregation of a relatively larger amount of global information.
\modify{
Dilated window attention mask is defined as
\begin{align} 
M_{ij} = \begin{cases}
0, & \text{if } |\phi(i) - \phi(j)| \% d = 0 , |\phi(i) - \phi(j)| \leq \dfrac{dw}{2} \\
-\infty, & \text{otherwise}
\end{cases}.
\end{align}
}
To further enhance the global information aggregation capability of LTCA, we introduce random attention, inspired by ShuffleNet~\cite{zhang2018shufflenet}. 
As shown in Fig.~\ref{fig:cpr_different_attn}(c), random attention enables each frame to attend to $r$ frames randomly selected from the entire video, which means that the receptive field of each frame is no longer limited, thereby enhancing globality.
Although random attention introduces randomness into inference, our experiments show that this randomness does not have a significant impact on inference (the standard deviation of metrics is less than 0.2\%).
\modify{
With $\psi_{\phi(i)}(T, r)$ denoting the randomly-selected $r$ frames from the set $\{1, 2, \ldots, T\}$ for $\phi(i)$, we define the random attention mask as
\begin{align}
M_{ij} = \begin{cases}
0, & \text{if } \phi(j) \in \psi_{\phi(i)}(T, r) \\
-\infty, & \text{otherwise}
\end{cases}.
\end{align}
}


\noindent
\textbf{Global Attention}
\modify{Although stacking sparse local attention can enable a global view of temporal context, sparse local attention may still not be able to fully capture the useful global context information.
Therefore, we further introduce global attention, which uses additionally introduced global queries to directly aggregate global context information.
We initialize the global queries with language features to regularize their training.}
As shown in Fig.~\ref{fig:cpr_different_attn}(b), global attention introduces $N_2$ additional queries $Q_g$, where all object queries $Q_o$ attend to global queries $Q_g$, and simultaneously, global queries $Q_g$ attend to the remaining object queries $Q_o$.
This operation ensures that any two queries can exchange information as long as two Transformer layers are stacked, solving the problem that the previous linear computational complexity attention could not fully model long videos. 
Global attention is a sparse attention mechanism, and its computational complexity is $O(2 N_2 N_1  T)$, which is still linear.
In global attention, the global queries aggregate $Q_g$ the contextual temporal information of each instance throughout the video, and the object queries of each frame attend to other objects through global queries as a bridge.
\modify{
Specifically, global attention mask can be expressed as
\begin{align}
M_{ij} = \begin{cases}
0, & \text{if } i\in \mathcal{G} \;\text{or}\; j\in \mathcal{G} \\
-\infty, & \text{otherwise}
\end{cases},
\end{align}
where $\mathcal{G}$ represents the global query set.
}

Furthermore, in LTCA, global queries $Q_g$ can be utilized to directly predict the object mask and confidence score as they have effectively gathered long-range temporal context information of each instance in the video with a good balance between locality and globality.
Therefore, the introduction of global attention negates the need for an additional video-object generator (\emph{i.e.,} Transformer Decoder in LMPM~\cite{ding2023mevis}) to construct global instance embeddings, further simplifying the framework for RVOS task.

\noindent
\textbf{Discussion}
\modify{
Sparse local attention (including dilated window attention and random attention) only attends to frames within a limited temporal range to maintain good locality. 
By stacking them layer by layer, a global view can be achieved. 
As sparse local attention mainly aims to enhance the locality, we design global queries to directly aggregate global context information.  
Thus, by combining sparse local attention and global attention, LTCA achieves a good balance between locality and globality.
}

\subsection{Mask Generator}
\label{sec:Mask_Generation}

\begin{table*}[!t]
\centering
\setlength{\tabcolsep}{11.5pt}
\caption{
Performance of our method on MeViS Datasets.
$^{\dagger}$ denotes the LMPM model retrained using our training parameters.
}
\begin{tabular}{l|c|c|ccc|ccc}
\toprule
\multirow{2}{*}{Methods}  &\multirow{2}{*}{Pub.\&Year} &\multirow{2}{*}{Backbone}                      & \multicolumn{3}{c|}{$val^u$}     & \multicolumn{3}{c}{$val$}        \\
 \cmidrule(lr){4-6} \cmidrule(lr){7-9}
                         &    &   & $\mathcal{J}\&\mathcal{F}$     & $\mathcal{J}$        & $\mathcal{F}$        & $\mathcal{J}\&\mathcal{F}$     & $\mathcal{J}$        & $\mathcal{F}$        \\
\midrule
URVOS~\cite{seo2020urvos}  &  ECCV 2020    & \multirow{8}{*}{Swin-T}     & -        & -        & -        & 27.8     & 25.7     & 29.9     \\
LBDT~\cite{ding2022language}   &  CVPR 2022     &                                          & -        & -        & -        & 29.3     & 27.8     & 30.8     \\
MTTR~\cite{botach2022end}  &   CVPR 2022    &                                                      & -        & -        & -        & 30.0     & 28.8     & 31.2     \\
ReferFormer~\cite{wu2022language}   &  CVPR 2022   &                                                & -        & -        & -        & 31.0     & 29.8     & 32.2     \\
VLT+TC~\cite{ding2022vlt}  &   TPAMI 2023    &                                                      & -        & -        & -        & 35.5     & 33.6     & 37.3     \\
LMPM~\cite{ding2023mevis}  &    ICCV 2023               &                                                       & 40.2     & 36.5     & 43.9     & 37.2     & 34.2     & 40.2     \\
LMPM$^{\dagger}$ & ICCV 2023 &  & 48.3 & 43.9 & 52.6 & 41.8 & 38.7 & 44.9  \\
\rowcolor[gray]{.9}
Ours                 &                          &                               & \textbf{51.5±0.1} & \textbf{46.6±0.2} & \textbf{56.4±0.1} & \textbf{45.3±0.1} & \textbf{41.7±0.1} & \textbf{48.9±0.1} \\
\midrule
\rowcolor[gray]{.9}
Ours            &    & Swin-S                                         & 52.9±0.2 & 47.8±0.2 & 58.1±0.3 & 45.9±0.1 & 42.0±0.1   & 49.7±0.0   \\
\midrule
\rowcolor[gray]{.9}
Ours         &       & Swin-B                                          & 55.4±0.1 & 50.8±0.1  &  60.0±0.1  & 46.6±0.1 & 43.2±0.1 & 50.0±0.1 \\
\midrule
\rowcolor[gray]{.9}
Ours        &        & Swin-L                                                  & 56.0±0.1   & 51.0±0.1   & 61.1±0.2 & 47.5±0.1 & 43.5±0.1 & 51.4±0.1 \\
\bottomrule
\end{tabular}
\label{tab:result_on_mevis}
\end{table*}

\begin{table*}
\centering
\setlength{\tabcolsep}{10.5pt}
\caption{
\modify{Performance of our method on Ref-YouTube-VOS and Ref-DAVIS17 Datasets.}
\modify{
Following conventional practice, 
``With Image Pretraining'' means the models are first pretrained on RefCOCO~\cite{kazemzadeh2014referitgame}, RefCOCO+~\cite{kazemzadeh2014referitgame}, and RefCOCOg~\cite{mao2016generation} datasets. 
``Joint Training'' means the models are trained with the combination of image datasets and video datasets.
}
}
\begin{tabular}{l|c|c|ccc|ccc}
\toprule
\multirow{2}{*}{Methods}&\multirow{2}{*}{Pub.\&Year} & \multirow{2}{*}{Backbone} & \multicolumn{3}{c|}{Ref-YouTube-VOS}   & \multicolumn{3}{c}{Ref-DAVIS17}    \\ 
\cmidrule(lr){4-6} \cmidrule(lr){7-9}
                         &    &                       & $\mathcal{J}\&\mathcal{F}$     & $\mathcal{J}$        & $\mathcal{F}$        & $\mathcal{J}\&\mathcal{F}$     & $\mathcal{J}$        & $\mathcal{F}$        \\
\midrule
\multicolumn{9}{c}{\textit{With Image Pretraining}}\\
\midrule
URVOS~\cite{seo2020urvos}    &  ECCV2020  &     \multirow{6}{*}{ResNet-50}                       & 47.2     & 45.3     & 49.2     & 51.5     & 47.3     & 56.0     \\
\modify{MLRL~\cite{wu2022multi}}  &  \modify{CVPR 2022}          &                  & \modify{49.7}     & \modify{48.4}     & \modify{51.0}     & \modify{52.7}     & \modify{50.1}     & \modify{55.4}  \\
ReferFormer~\cite{wu2022language}  &  CVPR 2022          &                   & 55.6     & 54.8     & 56.5     & 58.5     & 55.8     & 61.3  \\
\modify{OnlineRefer~\cite{wu2023onlinerefer}} & \modify{ICCV 2023}  &   & \modify{57.3} & \modify{55.6} & \modify{58.9} & \modify{59.3}  & \modify{55.7} & \modify{62.9} \\
\modify{TempCD~\cite{tang2023temporal}} & \modify{ICCV 2023}  &   & \modify{59.0} & \modify{57.5} & \modify{60.5} & \modify{60.0}  & \modify{57.3} & \modify{62.7}\\
\rowcolor[gray]{.9}
Ours          &            &                           & \textbf{59.7±0.2} & \textbf{58.1±0.1} & \textbf{61.3±0.2} & \textbf{60.6±0.1} & \textbf{57.4±0.1} & \textbf{63.7±0.1} \\
\midrule
ReferFormer~\cite{wu2022language}  &  CVPR 2022          &  \multirow{4}{*}{Video-Swin-T}                 & 59.4     & 58.0     & 60.9     & 59.7     & 56.6     & 62.8  \\
\modify{TempCD~\cite{tang2023temporal}} & \modify{ICCV 2023}  &   & \modify{62.3} & \modify{60.5} & \modify{64.0} & \modify{62.2}  & \modify{59.3} & \modify{65.0} \\
\modify{SOC~\cite{luo2024soc}}  &  \modify{NeurIPS 2023}       &                   & \modify{62.4}     & \modify{61.1}     & \modify{63.7}     & \modify{63.5}     & \modify{\textbf{60.2}}     & \modify{66.7}  \\
\rowcolor[gray]{.9}
Ours          &            &                           & \textbf{62.9±0.1} & \textbf{61.5±0.1} & \textbf{64.3±0.1} & \textbf{63.6±0.1} & 60.1±0.1 & \textbf{67.1±0.1} \\
\midrule
\multicolumn{9}{c}{\textit{Joint Training}}\\
\midrule
CMSA~\cite{ye2019cross}         &  CVPR 2019   & \multirow{8}{*}{ResNet-50}     & 34.9     & 33.3     & 36.5     & 34.7     & 32.2     & 37.2     \\
LBDT-4~\cite{ding2022language}  &   CVPR 2022    &                           & 48.2     & 50.6     & 49.4     & -        & -        & -        \\
YOFO~\cite{li2022you}       &  AAAI 2022   &                           & 48.6     & 47.5     & 49.7     & 53.3     & 48.8     & 57.9     \\
ReferFormer~\cite{wu2022language}  &  CVPR 2022          &                           & 58.7     & 57.4     & 60.1     & 61.1     & 58.0     & 64.1     \\
UniRef~\cite{wu2023segment} &  CVPR 2023  &                           & 60.6     & 59.0     & 62.3     & -        & -        & -        \\
MUTR~\cite{yan2023referred}  &      AAAI 2024             &                           & 61.9     & 60.4     & 63.4     & 65.3     & 62.4     & 68.2     \\
\rowcolor[gray]{.9}
Ours          &            &                           & \textbf{62.5±0.2} & \textbf{60.6±0.1} & \textbf{64.4±0.2} & \textbf{66.0±0.1} & \textbf{62.8±0.0} & \textbf{69.3±0.1} \\
\midrule
ReferFormer~\cite{wu2022language}    & CVPR 2022 & \multirow{4}{*}{Swin-L}    & 64.2     & 62.3     & 66.2     & 63.9     & 60.8     & 67.0     \\
UniRef~\cite{wu2023segment}     & CVPR 2023  &                           & 67.4     & 65.5     & 69.2     & -        & -        & -        \\
MUTR~\cite{yan2023referred}   &      AAAI 2024             &                           & 68.4     & 66.4     & 70.4     & 68.0     & 64.8     & 71.3     \\
\rowcolor[gray]{.9}
Ours          &            &                           & \textbf{69.1±0.1} & \textbf{67.1±0.1} & \textbf{71.2±0.0} & \textbf{68.6±0.1} & \textbf{65.2±0.1} & \textbf{71.9±0.1} \\
\bottomrule
\end{tabular}
\label{tab:result_on_refytvos_and_davis17}
\end{table*}

After the LTCA module, the global queries $\widetilde{Q}_{g}$ aggregate the temporal information of the entire video and represent several instances within the video. 
We feed the global queries and mask features into the Mask Generator to obtain the final predicted masks and confidence scores.
Specifically, the Mask Generator comprises of the segmentation head~$\mathcal{H}_{s}$ and the classification head~$\mathcal{H}_{c}$.
The segmentation head~$\mathcal{H}_{s}$ containing three MLP layers is applied to generate the set of segmentation masks~$\{M_i\}_{i=1}^{N_2}\in \mathbb{R}^{T\times N_2 \times \frac{H}{S} \times \frac{W}{S}}$, and each mask $M_i^t$ of the $t^{\text{th}}$ frame is obtained by

\begin{equation}
    M_{i}^{t} = F_m^t \otimes \mathcal{H}_{s}(\widetilde{Q}_{g_i}),
\end{equation}

\noindent
where the $\mathcal{H}_{s}(\widetilde{Q}_{g_i})$ indicates the dot product weight generated by global query of the last Transformer encoder layer, and the $\otimes$ means dot product operation. 
With the segmentation head, each global query generates segmentation masks of $T$ frames. 
Meanwhile, the global queries~$\widetilde{Q}_{g_i}$ are concatenated with the language feature $F_e$ and inputted into the classification head~$\mathcal{H}_{c}$, which consists of a single linear layer, to obtain the prediction confidence scores~$\{S_i\}_{i=1}^{N_2}\in \mathbb{R}^{N2}$ for prediction masks, which can be represented as follows:

\begin{equation}
    S_i = \mathcal{H}_{c}(\widetilde{Q}_{g_i}).
\end{equation}

Following LMPM~\cite{ding2023mevis}, we use $\mathcal{L}_{f}$ from \cite{cheng2022masked} to calculate loss from the per-frame outputs to frame-wise ground-truth, use $\mathcal{L}_{sim}$ and $\mathcal{L}_{v}$ from \cite{heo2022vita} to calculate the video-level loss. 
Finally, we integrate all losses together as follows:
\begin{equation}
    \mathcal{L}_{total} = \lambda_{v} \mathcal{L}_{v} + \lambda_{f}\mathcal{L}_{f} + \lambda_{sim}\mathcal{L}_{sim}.
\end{equation}

During the inference stage, for RVOS tasks that predict a single mask output, such as the Ref-YouTube-VOS dataset~\cite{seo2020urvos}, we select the mask with the highest prediction score as the final prediction result. 
For RVOS tasks that predict multiple mask outputs, such as the MeViS dataset~\cite{ding2023mevis}, we select the masks with prediction confidence scores greater than the confidence threshold $\sigma$ as the final prediction results.

\section{Experiments}

\subsection{Dataset and Evaluation Metrics}
The proposed methods are evaluated on \modify{four} video segmentation datasets: MeViS~\cite{ding2023mevis}, Ref-YouTube-VOS~\cite{seo2020urvos}, Ref-DAVIS17~\cite{khoreva2019video} and \modify{A2D-Sentences~\cite{gavrilyuk2018actor}}.
Following~\cite{ding2023mevis}, we use region similarity metric $\mathcal{J}$, F-measure $\mathcal{F}$ and the average
of these two metrics $\mathcal{J} \& \mathcal{F}$ as the evaluation metrics on MeViS Ref-YouTube-VOS and Ref-DAVIS17.
\modify{Following~\cite{botach2022end}, we use Overall IoU, Mean IoU and MAP over 0.50:0.05:0.95 as the evaluation metrics on A2D-Sentences.}
Following~\cite{wu2022language,yan2023referred}, on Ref-DAVIS17, we directly report the results using the model trained on Ref-YouTube-VOS without finetuning. 
For more detailed descriptions of the datasets and implementation details, please refer to the Appendix.

\begin{table}[!t]
\setlength{\tabcolsep}{3pt}
\centering
\caption{
\modify{Performance of our method on A2D-Sentences.}
}
\begin{tabular}{l|c|c|cc|c}
\toprule
\multirow{2}{*}{Method} & \multirow{2}{*}{Pub.\&Year} & \multirow{2}{*}{Backbone} & \multicolumn{2}{c|}{IoU} & \multirow{2}{*}{mAP} \\ \cmidrule(lr){4-5}
                        &                      &                           & Overall      & Mean     &                      \\
\midrule
\modify{ACAN~\cite{wang2019asymmetric}}                    & \modify{ICCV 2019}            & \modify{I3D}                       & \modify{60.1}         & \modify{49.0}     & \modify{27.4}                \\
\modify{CMPC-V~\cite{liu2021cross} }                & \modify{TPAMI 2021}           & \modify{I3D}                       & \modify{65.3}         & \modify{57.3}     & \modify{40.4}                 \\
MTTR~\cite{botach2022end}                   & CVPR 2022            & Video-Swin-T              & 72.0         & 64.0     & 46.1                 \\
ReferFormer~\cite{wu2022language}            & CVPR 2022            & Video-Swin-T              & 72.3         & 64.1     & 48.6                 \\
SOC~\cite{luo2024soc}                    & NeurIPS 2023         & Video-Swin-T              & 74.7         & 66.9     & 50.4                 \\
\rowcolor[gray]{.9}
Ours                    &                      & Video-Swin-T              & \textbf{75.5}         & \textbf{67.7}     & \textbf{51.4}                 \\
\bottomrule
\end{tabular}
\label{tab:result_on_a2d}
\end{table}

\subsection{Comparison with previous state-of-the-arts}
We compare our method to previous state-of-the-art RVOS methods in Table~\ref{tab:result_on_mevis} and Table~\ref{tab:result_on_refytvos_and_davis17}.
The numbers reported in the tables are directly cited from corresponding papers, except for LMPM$^{\dagger}$ which is retrained using our training setups.

\noindent \textbf{MeViS}
We compare our method to previous works on MeViS benchmark in Table~\ref{tab:result_on_mevis}.
Our approach outperforms all the previous works remarkably, \emph{e.g.,} outperforming LMPM by around $11.3\%$ and $8.1\%$ $\mathcal{J} \& \mathcal{F}$ on MeViS $val^u$ and $val$ datasets. As MeViS contains more motion expressions than the other two benchmarks, the superior performance of our method verifies the effectiveness of our LTCA in aggregating temporal information.

\noindent
\textbf{Ref-YouTube-VOS \& Ref-DAVIS17}
We compare our method to previous models on Ref-YouTubeVOS and Ref-DAVIS17 in Table~\ref{tab:result_on_refytvos_and_davis17}. 
\modify{Our method achieves new state-of-the-art performance among different training settings: with image pretraining, and joint training.}
For example, our method outperforms MUTR by about $0.7\%$ $\mathcal{J} \& \mathcal{F}$ on Ref-YouTube-VOS dataset with Swin-L backbone.
The less significant improvement on Ref-YouTube-VOS and Ref-DAVIS17 may be attributed to the fact that these datasets primarily focus on static objects with fewer language expressions depicting motion.
In contrast, our method concentrates on long-range temporal context modeling.

\noindent
\modify{\textbf{A2D-Sentences}
As shown in Table~\ref{tab:result_on_a2d}, our method outperforms the current state-of-the-art methods, SoC~\cite{luo2024soc} and ReferFormer~\cite{wu2022language}. When trained from scratch, our model achieves 51.4\% mAP and 67.7\% mIoU, with an improvement of 1.0\% mAP and 0.8\% mIoU compared to SOC.}

\subsection{Ablation Studies}

\noindent \textbf{Effect of Each Component.} 
In Table~\ref{tab:eff_component}, we conduct experiments to verify the effectiveness of each key component in our LTCA.
When combining the global attention with the shift window attention (adopted by LMPM), the performance improves $2.2\%$ $\mathcal{J} \& \mathcal{F}$ on the $val^u$ set compared with baseline model LMPM$^{\dagger}$, which verifies the effectiveness of our proposed global attention.
By comparing the first two rows of the table, we observe that introducing the global attention may avoid using the Video-Object Generator.
From the last three rows of the table, replacing the dense local attention (\emph{i.e.,} shift window attention) with each kind of sparse local attention (\emph{i.e.,} dilated window attention and random attention) yields an obvious improvement, verifying the effectiveness of our proposed sparse local attention. 
Both the global attention and the sparse local attention contribute to the final video object segmentation performance.

\begin{table*}
\centering
\setlength{\tabcolsep}{9pt}
\caption{\modify{Effect of each attention component of LTCA on MeViS.}}
\label{tab:eff_component}
\begin{tabular}{cccc|ccc|ccc}
\toprule
\multirow{2}{*}{Global Attention} & \multicolumn{2}{c}{Sparse Local Attention}        & Video-Object & \multicolumn{3}{c|}{$val^u$}       & \multicolumn{3}{c}{$val$}        \\
\cmidrule(lr){2-3} \cmidrule(lr){5-7} \cmidrule(lr){8-10}
                              & Dilated & Random & Generator    & $\mathcal{J}\&\mathcal{F}$     & $\mathcal{J}$        & $\mathcal{F}$       & $\mathcal{J}\&\mathcal{F}$     & $\mathcal{J}$        & $\mathcal{F}$       \\
\midrule
\ding{51}                                  &                      &              &  \ding{51} & 50.5     & 45.9     & 55.1     & 42.8     & 39.5     & 46.0     \\
\ding{51}                                 &                      &              & \ding{56}            & 50.6     & 45.5     & 55.6     & 44.0     & 40.3     & 47.8     \\
\modify{\ding{51}}                              &        \modify{\ding{51}}               &              & \modify{\ding{56}}            & \modify{51.0}     & \modify{45.9}     & \modify{56.1}     & \modify{44.3}     & \modify{40.4}     & \modify{48.1}     \\
\ding{51}                                &                      & \ding{51}            & \ding{56}            & 51.2±0.1 & 46.3±0.1 & 56.0±0.0 & 44.8±0.0 & 41.2±0.0 & 48.5±0.0 \\
\rowcolor[gray]{.9}
\ding{51}                                & \ding{51}                    & \ding{51}            & \ding{56}            & 51.5±0.1 & 46.6±0.2 & 56.4±0.1 & 45.3±0.1 & 41.7±0.1 & 48.9±0.1 \\
\bottomrule
\end{tabular}

\end{table*}
\begin{table*}
    \setlength{\tabcolsep}{3.6pt}
    \caption{Comparison of metrics on long videos and short videos on the MeViS dataset. 
$\Delta \mathcal{J}\&\mathcal{F}$ represents the percentage change of metrics on long videos relative to short videos.}
\hspace{8pt}
    \begin{minipage}{0.46\textwidth} 
        \centering
        \caption*{  (a) We divided the MeViS $val$ set into short videos and long videos, trying to equalize the number of long videos and short videos. \\ }

\begin{tabular}{l|ccc|ccc|c}
\toprule
\multirow{2}{*}{Method} & \multicolumn{3}{c|}{Video}          & \multicolumn{3}{c|}{$val$} & $\Delta \mathcal{J}\&\mathcal{F}$  \\
\cmidrule(lr){2-4} \cmidrule(lr){5-7}
                        & Type   & Length            & Number & $\mathcal{J}\&\mathcal{F}$ & $\mathcal{J}$ & $\mathcal{F}$  & (relative)   \\
\midrule
\multirow{2}{*}{LMPM}   & Short  & $\leq 46$   & 1121   & 38.0   & 34.9   & 41.0  & \multirow{2}{*}{\red{-4.1\%}} \\
                        & Long & $>46$ & 1115   & 36.4   & 33.5   & 39.3  &                         \\
\midrule
\multirow{2}{*}{LMPM$^{\dagger}$}  & Short  & $\leq 46$   & 1121   & 41.9   & 38.9   & 44.9  & \multirow{2}{*}{\red{-0.5\%}} \\
                        & Long & $>46$ & 1115   & 41.7   & 38.5   & 44.9  &                         \\
\midrule
\multirow{2}{*}{Ours}    & Short  & $\leq 46$   & 1121   & 44.9   & 41.5   & 48.3  & \multirow{2}{*}{\green{+2.2\%}}  \\
                        & Long & $>46$ & 1115   & 45.9   & 42.0   & 49.7  &      \\
\bottomrule
\end{tabular}

    \end{minipage}
    \hspace{10pt}
    \begin{minipage}{0.46\textwidth} 
        \centering
        \caption*{(b) We selected videos where information exchange can occur between any two frames under the LMPM shift window attention mechanism as short videos, and the rest as long videos, with the video threshold set at 18.}

\begin{tabular}{l|ccc|ccc|ccc}
\toprule
\multirow{2}{*}{Method} & \multicolumn{3}{c|}{Video}          & \multicolumn{3}{c|}{$val$} & $\Delta \mathcal{J}\&\mathcal{F}$  \\
\cmidrule(lr){2-4} \cmidrule(lr){5-7}
                        & Type   & Length            & Number & $\mathcal{J}\&\mathcal{F}$ & $\mathcal{J}$ & $\mathcal{F}$  & (relative)   \\
\midrule
\multirow{2}{*}{LMPM}   & Short & $\leq 18$   & 21     & 56.9   & 55.5   & 58.3  & \multirow{2}{*}{\red{-34.9\%}}  \\
                        & Long  & $> 18$ & 2215   & 37.0   & 34.0   & 40.0  &                                    \\
\midrule
\multirow{2}{*}{LMPM$^{\dagger}$}  & Short & $\leq 18$   & 21     & 63.9   & 62.6   & 65.2  & \multirow{2}{*}{\red{-34.9\%}} \\
                        & Long  & $> 18$ & 2215   & 41.6   & 38.5   & 44.7  &                                     \\
\midrule
\multirow{2}{*}{Ours}    & Short & $\leq 18$   & 21     & 57.8   & 56.3   & 59.3  & \multirow{2}{*}{\green{-21.7\%}} \\
                        & Long  & $> 18$ & 2215   & 45.3   & 41.6   & 48.9  &                             \\
\bottomrule
\end{tabular}

    \end{minipage}
    \label{tab:ablation_on_video_len}

\end{table*}

\noindent \textbf{Effect of Video Length.}
\label{ablation_study_on_video_length}
a)
In Table~\ref{tab:ablation_on_video_len}(a), we split the MeViS $val$ set into short and long video groups based on the number of video frames. 
The results show that our method, performs better on long videos than on short ones, with an absolute advantage of 1\% and a relative advantage of 2.2\% on $\mathcal{J}\&\mathcal{F}$. 
In contrast, LMPM performs relatively worse on long videos than short videos. 
The comparisons show that compared to LMPM which stacks dense local attentions (\emph{i.e.,} shift window attention), our LTCA is more effective to aggregate long-range temporal information. 

\noindent
b) 
As we know, LMPM ensures a global view by staking multiple shift window attention layers.
However, with a certain number of layers, only partial frames can exchange their information, \emph{i.e.,} the temporal receptive field $L$ of each frame in the last layer relative to the first layer may be smaller than the video length.
We use $L$ to separate short and long videos, \emph{i.e.,} the videos whose length is larger than $L$ are treated as long videos, and those whose length is smaller than $L$ are viewed as short videos.
The results in Table~\ref{tab:ablation_on_video_len}(b) show that under this division method, the performance loss of our method on long videos relative to short videos is lower than that of LMPM, further validating the superior ability of our method to model long videos. 
The reason we use sample quantity as the basis for division in Table~\ref{tab:ablation_on_video_len}(a) is that text descriptions with video frames less than or equal to 18 in the MeViS val set only account for 1\% of the $val$ set, which may lead to less credible results. 
However, the comparisons under both division methods show that our method is superior to LMPM in modeling long videos.

\noindent \textbf{Effect of Sparse Local Attention hyper-parameters.}
In Table~\ref{tab:ablation_of_LTCA}, we conduct experiments on various hyper-parameters in sparse local attention of LTCA, including the number of attending neighbors $w$, dilated interval $d$ and the number of randomly selected frames $r$ respectively. 
From the table, we observe that the optimal choices for $w$, $r$ and $d$ should neither be too large nor too small.
We speculate it is because the optimal choices reach a good balance between locality and globality.

\noindent
\textbf{Comparison with Full Attention}
\modify{
We firstly replace the sparse local attention in LTCA with full attention and train the model on MeViS.
According to results in Table~\ref{tab:result_on_full_with_ltca}, training with dense global attention (Line 1) is inferior than LTCA (Line 3). 
Furthermore, we train a model with a combination of full attention and LTCA, which means a layer-wise alternation between the two attention mechanisms within the transformer encoder~(LTCA module).
We observe that such a combination (Line 2) also results in inferior results compared with LTCA.  
}

\begin{table}[!t]
\centering
\setlength{\tabcolsep}{9.5pt}
\caption{
Effect of different sparse local attention setups on MeViS. 
The $w$, $r$ and $d$ are determined sequentially.
}
\begin{tabular}{ccc|ccc}
\toprule
\multirow{2}{*}{\; $w$\; } & \multirow{2}{*}{\; $r$\; } & \multirow{2}{*}{\; $d$\; } & \multicolumn{3}{c}{$val$}        \\ \cmidrule(lr){4-6}
 &  &  & $\mathcal{J}\&\mathcal{F}$ & $\mathcal{J}$ & $\mathcal{F}$   \\
\midrule
1 & 0 & 1 & 43.5     & 39.6     & 47.4     \\
3 & 0 & 1 & \textbf{44.0} & \textbf{40.3} & \textbf{47.8}\\
5 & 0 & 1 & 43.9     & 40.4     & 47.3     \\ \midrule
3 & 0 & 1 & 44.0     & 40.3    & 47.8    \\
3 & 1 & 1 & 44.3±0.1 & 40.5±0.1 & 48.1±0.0 \\
3 & 2 & 1 & \textbf{44.8±0.0} & \textbf{41.2±0.0} & \textbf{48.5±0.0} \\
3 & 3 & 1 & 44.6±0.2 & 40.9±0.2 & 48.3±0.1 \\ \midrule
3 & 2 & 1 & 44.8±0.0 & 41.2±0.0 & 48.5±0.0 \\
\rowcolor[gray]{.9}
3 & 2 & 2 & \textbf{45.3±0.1} & \textbf{41.7±0.1} & \textbf{48.9±0.1} \\
3 & 2 & 3 & 44.1±0.1 & 40.5±0.1 & 47.7±0.1 \\
\bottomrule
\end{tabular}

\label{tab:ablation_of_LTCA}

\end{table}
\begin{table}[!t]
\centering
\caption{
\modify{Comparison of LTCA and full attention on the MeViS dataset.}
}
\begin{tabular}{l|ccc|ccc}
\toprule
\multirow{2}{*}{Method} & \multicolumn{3}{c|}{$val^u$} & \multicolumn{3}{c}{$val$}    \\ \cmidrule(lr){2-4} \cmidrule(lr){5-7}
                        & $\mathcal{J}\&\mathcal{F}$     & $\mathcal{J}$        & $\mathcal{F}$   & $\mathcal{J}\&\mathcal{F}$     & $\mathcal{J}$        & $\mathcal{F}$      \\
\midrule
Full Attn.              & 50.8     & 45.8     & 55.7   & 42.6& 39.8&45.4  \\
Full Attn. + LTCA       & 51.1 & 46.2 & 56.0  & 44.3&40.8&47.8 \\
\rowcolor[gray]{.9}
LTCA                    & 51.5 & 46.6 & 56.4 &45.3 & 41.7 & 48.9\\
\bottomrule
\end{tabular}
\label{tab:result_on_full_with_ltca}
\end{table}
\begin{table}[!t]
\setlength{\tabcolsep}{5.6pt}
\centering
\caption{
\textbf{Efficiency comparison with state-of-the-art methods.} 
The shape of the input videos is [48, 3, 448, 796], and the length of input text tokens is 40. 
}
\begin{tabular}{l|cccc}
\toprule
Methods     & GFLOPs & Parameters & Infer. Time & GPU Memory \\
\midrule
ReferFormer & 7172 & \textbf{176.4M} & 1.39s       & 25540M     \\
MUTR        & 7220 & 186.4M & 1.42s       & 25641M     \\
LMPM        & \underline{4430} & 191.0M & \underline{0.98s}       & \underline{8165M}      \\
\rowcolor[gray]{.9}
Ours         & \textbf{4429} & \underline{186.3M} & \textbf{0.97s}       & \textbf{8121M}    \\
\bottomrule
\end{tabular}

\label{tab:effectiveness_of_zoe}




\end{table}


\begin{table}[t]
\setlength{\tabcolsep}{9pt}
\caption{\modify{Performance comparison with varying video lengths on the MeViS dataset.} }
\centering
\begin{tabular}{cc|ccc}
\toprule
\multicolumn{2}{c|}{Video} & \multicolumn{3}{c}{$val$}        \\
\cmidrule(lr){1-2}\cmidrule(lr){3-5}
Length         & Number   & $\mathcal{J}\&\mathcal{F}$     & $\mathcal{J}$        & $\mathcal{F}$        \\
\midrule
$15 \leq L < 35$      &35      & 43.7±0.2 & 40.9±0.2 & 46.5±0.2 \\
$35 \leq L < 48$         & 35       & 45.7±0.1 & 41.8±0.1 & 49.5±0.1 \\
$48 \leq L < 79$       & 35       & 45.9±0.1 & 42.9±0.1 & 48.9±0.1 \\
$79 \leq L < 293$      & 35       & 45.6±0.0 & 40.7±0.0 & 50.5±0.0 \\
\rowcolor[gray]{.9}
ALL            & 140      & 45.3±0.1 & 41.7±0.1 & 48.9±0.1 \\
\bottomrule
\end{tabular}
\label{tab:random_at_different_length}
\end{table}
\begin{table}[!t]
\centering
\caption{Metrics of our model on the MeViS dataset under different random seed settings.}
\begin{tabular}{c|ccc|ccc}
\toprule
\multirow{2}{*}{Random Seed} & \multicolumn{3}{c|}{$val^u$}                                               & \multicolumn{3}{c}{$val$}                                                  \\ \cmidrule(lr){2-4} \cmidrule(lr){5-7}
                             & \multicolumn{1}{c}{$\mathcal{J}\&\mathcal{F}$} & \multicolumn{1}{c}{$\mathcal{J}$} & \multicolumn{1}{c|}{$\mathcal{F}$} & \multicolumn{1}{c}{$\mathcal{J}\&\mathcal{F}$} & \multicolumn{1}{c}{$\mathcal{J}$} & \multicolumn{1}{c}{$\mathcal{F}$} \\
\midrule
0                            & 51.4                     & 46.5                  & 56.3                  & 45.4                     & 41.8                  & 49.0                  \\
1                            & 51.6                     & 46.7                  & 56.5                  & 45.2                     & 41.6                  & 48.8                  \\
2                            & 51.7                     & 46.8                  & 56.6                  & 45.3                     & 41.7                  & 48.9                  \\
3                            & 51.4                     & 46.5                  & 56.2                  & 45.4                     & 41.7                  & 49.0                  \\
4                            & 51.7                     & 46.8                  & 56.5                  & 45.3                     & 41.7                  & 49.0                  \\
5                            & 51.6                     & 46.7                  & 56.5                  & 45.4                     & 41.8                  & 49.0                  \\
6                            & 51.5                     & 46.6                  & 56.3                  & 45.4                     & 41.7                  & 49.0                  \\
7                            & 51.6                     & 46.7                  & 56.5                  & 45.3                     & 41.6                  & 48.9                  \\
8                            & 51.5                     & 46.6                  & 56.4                  & 45.4                     & 41.8                  & 49.0                  \\
\bottomrule
\end{tabular}
\label{tab:Robustness_of_LTCA}
\end{table}

\begin{figure}[!t]
  \centering
  \caption{
\textbf{Visualization Results on MeViS.}
We compare the visualization results between LMPM and our method on MeViS.
Previous work easily segments irrelevant objects, as this method fails to strike a good balance between locality and globality.
}
\vspace{-0.9mm}
  \includegraphics[width=0.99\linewidth]{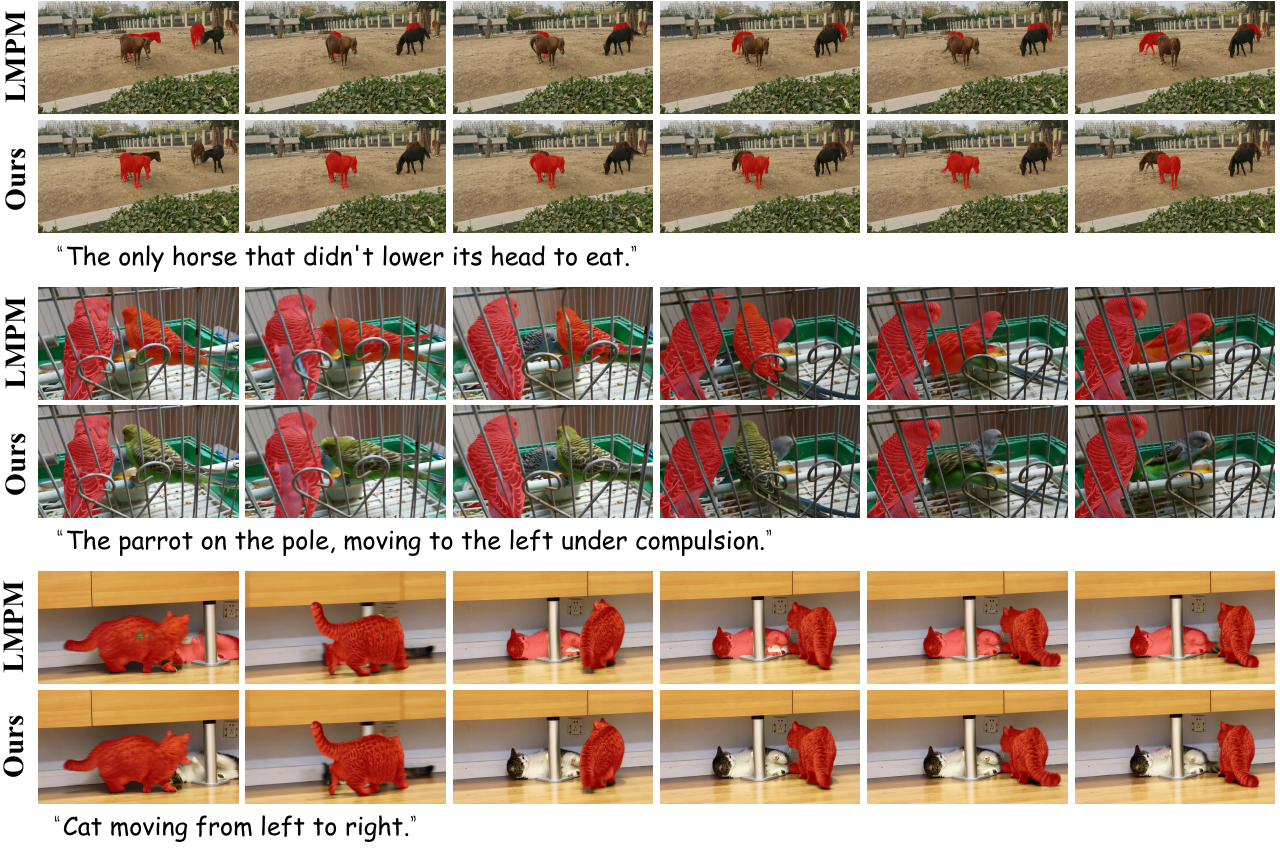}
\label{fig:visualization}
\vspace{-2.5mm}
\end{figure}

\begin{figure}[!t]
  \centering
\caption{
\textbf{Visualization Results on Ref-YouTube-VOS.}
We compare the visualization results between MUTR and our method on Ref-YouTube-VOS benchmarks.
}
\vspace{-0.9mm}
\label{fig:more_vis_refytvos}
  \includegraphics[width=0.99\linewidth]{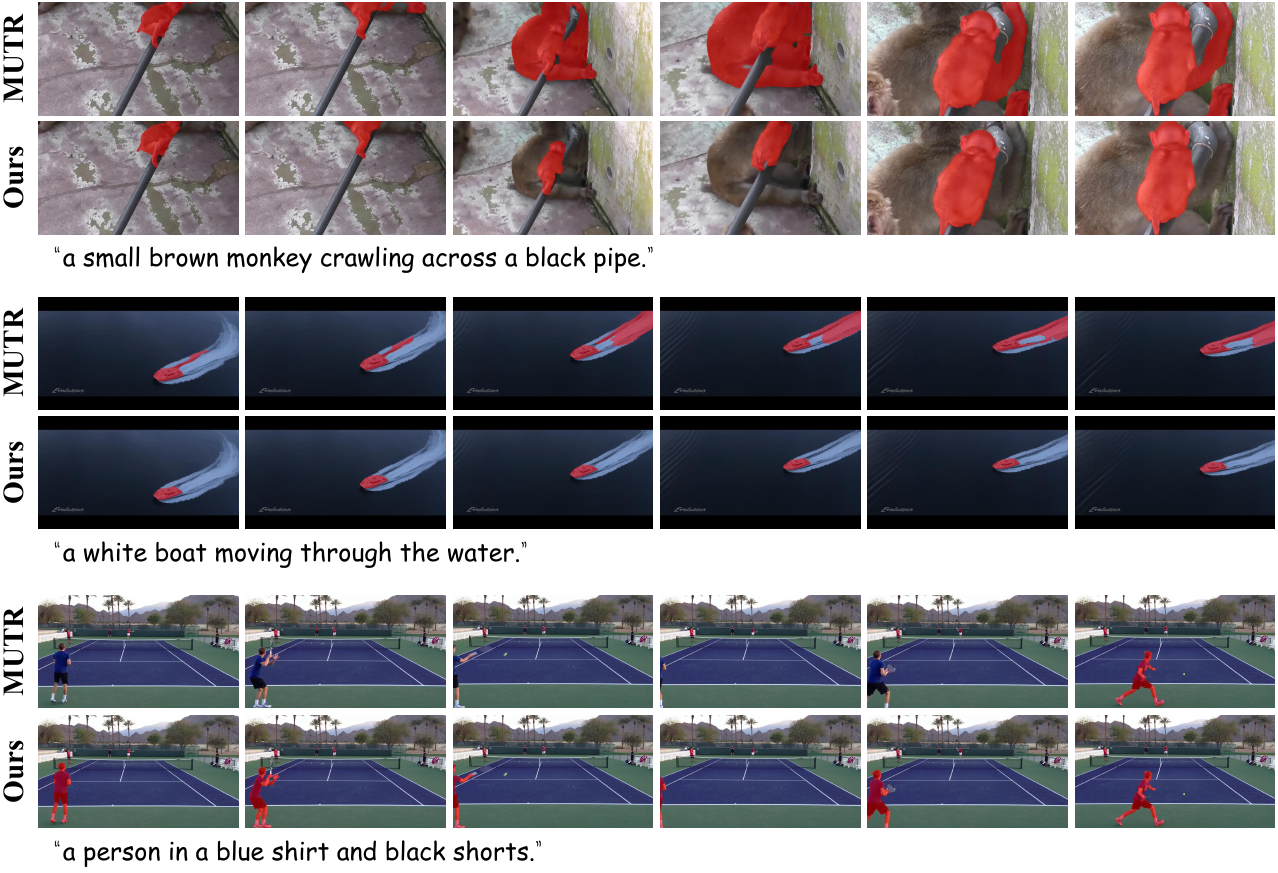}
\vspace{-2.5mm}
\end{figure}

\begin{figure}[!t]
  \centering
\caption{
\modify{\textbf{Failure Case of LTCA on MeViS.} 
Frames where the target of the textual description appears are marked in blue.
The segmentation results are shown with a red mask. The objects within the yellow boxes represent the ground-truth targets.
}
}
\label{fig:failure_case}
  \includegraphics[width=0.99\linewidth]{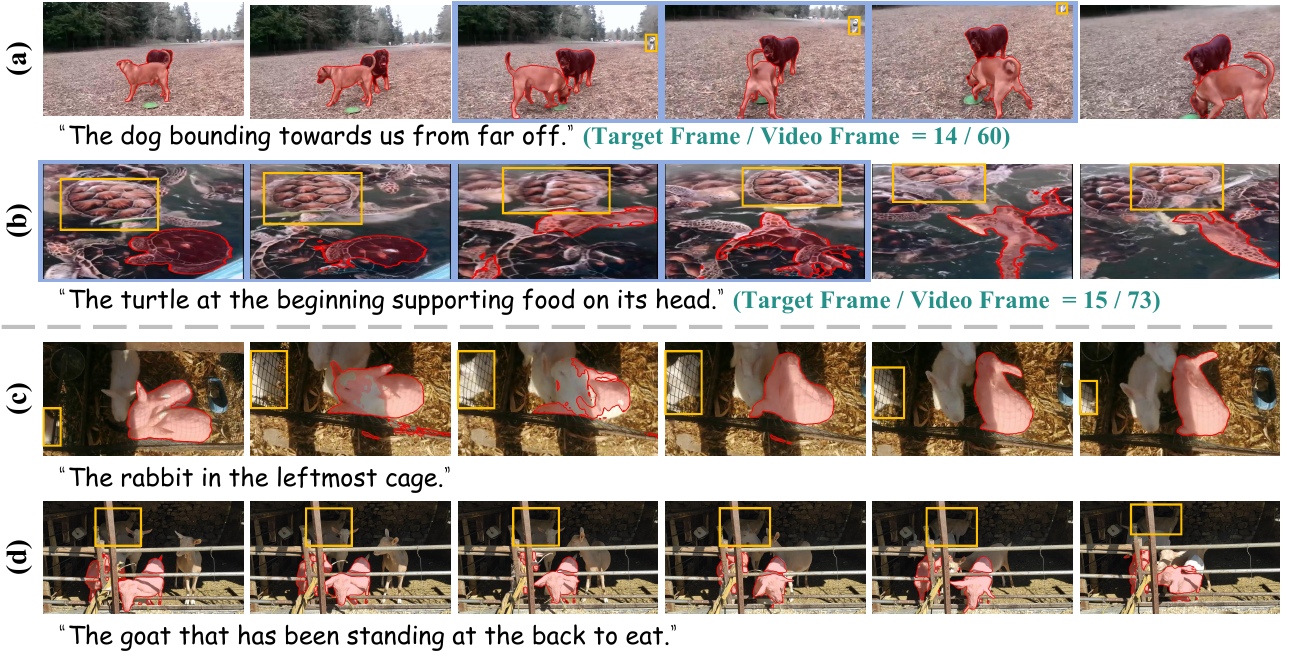}
\end{figure}

\noindent \textbf{Efficiency of Our Method.} In Table~\ref{tab:effectiveness_of_zoe}, we compare FLOPs, parameters, inference speed and GPU memory of our method with ReferFormer, MUTR and LMPM. 
As we avoid using a video-object generator to generate masks across frames, 
the number of parameters in our model is slightly lower than that of LMPM.
Our method exhibits a computational complexity similar to that of LMPM, as both our proposed LTCA and the shifted window attention mechanism utilized by LMPM maintain a linear complexity in terms of GLOPs.

\noindent
\textbf{Visualizations.}
Fig.~\ref{fig:visualization} shows qualitative results of both our method and LMPM.
While LMPM is prone to segmenting wrong objects, ours in contrast is more accurate in localizing the target objects according to the text.
This improvement can be attributed to our proposed LTCA which provides superior long-range temporal context modeling for videos.
In Fig.~\ref{fig:more_vis_refytvos}, we visualize the results compared with MUTR on Ref-YouTube-VOS benchmark. 
From Fig.~\ref{fig:more_vis_refytvos}, our method can successfully track and segment the referred instance even in challenging situations.

\noindent \textbf{Robustness of LTCA's Random Attention.}
Random Attention in LTCA allows object embeddings of each frame to attend to the embeddings of other frames randomly, which introduces randomness into the inference stage. 
\modify{The results presented in Tables~\ref{tab:result_on_mevis},~\ref{tab:result_on_refytvos_and_davis17}, and~\ref{tab:result_on_a2d} validate the robustness of our method across datasets of varying complexity.}
We conducted the following ablations to \modify{further} verify the robustness of our proposed random attention.
\modify{
Firstly, we validate LTCA on MeViS under different random seed settings in Table~\ref{tab:Robustness_of_LTCA}, which demonstrates that our proposed LTCA is robust.
We further validate the stability of random attention under different video lengths (Table~\ref{tab:random_at_different_length}).
Based on the results presented above, we note that all experiments exhibit quite small standard deviations (approximately 0.1\%), validating the stability of random attention under different video lengths and video complexities. 
}

\subsection{Failure Case Study}

\modify{
We observe that LTCA may not perform well in the following two situations: (1) dealing with long videos containing scarce frames of target object (as shown in Fig.~\ref{fig:failure_case} (a,~b)); (2) inferring relative spatial positions (as shown in Fig.~\ref{fig:failure_case} (c,~d)).
In future, we may pre-estimate the text-related frames before aggregating temporal context information and utilize Large Language Models (LLMs) to enhance the understanding of relative spatial positions to deal with the above two limitations respectively.}

\section{Conclusion}

In this paper, we propose a novel attention mechanism, LTCA, designed to capture the long-range temporal context for RVOS. LTCA employs a stack of sparse local attention layers, including dilated window attention and random attention, to strike a balance between locality and globality. It also directly aggregates global information through specifically designed global queries.
Our simplified framework achieves new state-of-the-art performance on \modify{four} RVOS benchmarks compared to previous solutions. We hope that our proposed LTCA and simplified framework may inspire future studies on long video-related tasks. 

\vspace{-1mm}
\bibliographystyle{IEEEtran}
\bibliography{reference}

\end{document}